\pdfoutput=1

\documentclass[11pt]{article}

\usepackage[]{emnlp2021}

\usepackage{times}
\usepackage{latexsym}

\usepackage[T1]{fontenc}

\usepackage[utf8]{inputenc}

\usepackage{microtype}
\usepackage{multirow}
\usepackage{booktabs}
\usepackage{graphicx}
\usepackage{mylingmacros}
\usepackage{xcolor}
\newcommand{\conn[1]}{\underline{#1}}
\newcommand{\argi[1]}{\textit{#1}}
\newcommand{\argii[1]}{\textbf{#1}}
\title{Revisiting Shallow Discourse Parsing in the PDTB-3: Handling Intra-sentential Implicits}

\author{Zheng Zhao \and Bonnie Webber \\
        Institute for Language, Cognition and Computation \\
      School of Informatics, University of Edinburgh \\
      10 Crichton Street, Edinburgh, EH8 9AB \\
      \texttt{\{zheng.zhao,bonnie.webber\}@ed.ac.uk}}

\begin{document}
\maketitle
\begin{abstract}
In the PDTB-3, several thousand implicit discourse relations were newly annotated \textit{within} individual sentences, adding to the over 15,000 implicit relations annotated \textit{across} adjacent sentences in the PDTB-2. Given that the position of the arguments to these \textit{intra-sentential implicits} is no longer as well-defined as with \textit{inter-sentential implicits}, a discourse parser must identify both their location and their sense.
That is the focus of the current work. The paper provides a comprehensive analysis of our results, showcasing model performance under different scenarios, pointing out limitations and noting future directions. 
\end{abstract}

\section{Introduction}
Discourse parsing is the task of identifying and categorizing discourse relations between discourse segments in a given text. The task is considered to be important for downstream tasks such as question answering \cite{jansen-etal-2014-discourse}, machine translation \cite{li-etal-2014-assessing}, and text summarization \cite{cohan-etal-2018-discourse}.
There are various approaches to discourse parsing, corresponding to different views of (1) what constitutes the segments of discourse, (2) what structures can be built from such segments, and (3) what semantic and/or rhetorical relations can hold between such segments \cite{xue-etal-2015-conll,zeldes-etal-2019-disrpt}.

Approaches to discourse structure generally allow discourse relations to hold between segments within a sentence (i.e., \textit{intra-sentential discourse relations}) or across sentences (i.e. \textit{inter-sentential discourse relations}) \cite{joty-etal-2012-novel,muller-etal-2012-constrained,stede2011discourse,stede-etal-2016-parallel}.

In the Penn Discourse Treebank (PDTB; \citealp{prasad-etal-2008-penn}), all discourse relations have two arguments, called \argi[Arg1] and \argii[Arg2]. Discourse relations are termed \textit{explicit}, if the evidence for the relation is an explicit discourse connective (word or phrase). For \textit{implicit discourse relations}, evidence is in the form of argument adjacency (with or without intervening punctuation), though annotators were asked to record one or more discourse connectives that, if present, would explicitly signal the sense(s) they inferred to hold between the arguments. Where annotators felt that the relation was already signalled by an alternative (non-connective) expression, the expression was annotated as evidence for what was called an \textit{AltLex relation} \cite{prasad-etal-2010-realization}.

The first major release of the PDTB was the PDTB-2
\cite{prasad-etal-2008-penn} 
whose guidelines limited annotation to (a) Explicit relations lexicalized by discourse connectives, and (b) Implicit and AltLex
relations between paragraph-internal adjacent sentences and between complete clauses within sentences separated by colons or semi-colons.
Since there were only $\sim$530 intra-sentential implicit relations among the $\sim$15,500 implicit relations annotated in the PDTB-2, they were ignored in work on discourse parsing \cite{lin2014pdtb,wang-lan-2015-refined,xue-etal-2015-conll,xue-etal-2016-conll}, which took implicit relations to hold only between adjacent sentences.

The situation changed with the release of the PDTB-3 \cite{pdtb3-manual:2019}. Among the $\sim$5.6K sentence-internal implicit relations annotated in the PDTB-3
are relations between VPs or clauses conjoined implicitly by punctuation (Ex.~\ref{ex:demon1}), between a free adjunct or free to-infinitive and its matrix clause (Ex.~\ref{ex:demon2}), and between a marked syntactic construction and its matrix clause. There are also implicit relations co-occurring with explicit relations \citep{pdtb3-manual:2019}, as noted in Section~\ref{sec:dataset}.

\enumsentence{\label{ex:demon1}
\small
Father McKenna moves through the house \argi[praying in Latin], \conn[(Implicit=and)] \argii[urging the demon to split]. [wsj\_0413] 
}

\enumsentence{\label{ex:demon2}
\small
\argi[Father McKenna moves through the house] \conn[(Implicit=while)] \argii[praying in Latin], \argii[urging the demon to split]. [wsj\_0413] 
}

Why are implicit \textit{intra-sentential} relations a problem for shallow discourse parsing? First, as already noted, unlike \textit{inter-sentential} implicits, they do not occur at sentence boundaries, with material to the left of the boundary as \argi[Arg1] and material to the right as \argii[Arg2]. Secondly, \argi[Arg1] and \argii[Arg2]
can appear in either order: \argi[Arg1] before \argii[Arg2], as in Ex~\ref{ex:demon1}--\ref{ex:demon2}, or \argii[Arg2] before \argi[Arg1], as in Ex.~\ref{ex:arg2-arg1}. Parsing implicit intra-sentential relations therefore requires both locating and labelling their arguments, as well as identifying the sense(s) in which they are related.\footnote{While the arguments to some explicit discourse relations can also appear in either order, \argii[Arg2] is still bound to the explicit conjunction, so its relative position can be easily identified.}

\enumsentence{\label{ex:arg2-arg1}
\small
\conn[(Implicit=if it is)] \argii[To slow the rise in total spending], \argi[it will be necessary to reduce per-capita use of services].
[wsj\_0314]}


This work takes up some of the challenges of parsing implicit intra-sentential discourse relations.
Overall, it contributes: (1) a set of BERT-based models used as a pipeline for recognizing intra-sentential implicit discourse relations as well as classifying their senses; (2) experimental evidence that these BERT-based models perform better than comparable LSTM-based models on the relation recognition task; 
(3) evidence that the use of parse tree features can improve model performance, as was earlier found useful in simply recognizing when a sentence contained at least one implicit intra-sentential relation \cite{liang-etal-2020-extending}.



\section{Related Work}

The focus of the current work is parsing implicit intra-sentential discourse relations in the framework of the PDTB-3. As most of the implicit relations in the PDTB-2 were inter-sentential (i.e., $\sim$95\% of its 15.5K implicit relations), its intra-sentential implicits were ignored in parser development. Nearly all recent work on recognizing inter-sentential implicits in the PDTB-2 used neural architectures. This included multi-level attention
in work by \citet{liu-li-2016-recognizing}, multiple text representations in work by \citet{bai-zhao-2018-deep}, including character, subword, word, sentence, and sentence pair levels, to more fully capture the text. \citet{dai-huang-2018-improving} introduced a paragraph-level neural architecture with a conditional random field (CRF, \citealp{LaffertyMP01}) layer which models inter-dependencies of discourse units and predicts a sequence of discourse relations in a paragraph. \citet{varia-etal-2019-discourse} introduced an approach to distill knowledge from word pairs for discourse relation with CNN by joint learning of implicit and explicit relations. \citet{shi-demberg-2019-next} discovered that BERT-based models, which were trained on the next sentence prediction task, benefited implicit inter-sentential discourse relation classification. Here we assess whether they also benefit classifying intra-sentential implicit relations. 

Looking at implicit relations in the PDTB-3, \citet{prasad-etal-2017-towards} consider the difficulty in extending implicit relations to relations that cross paragraph boundaries. \citet{kurfali-ostling-2019-zero} examine whether implicit relation annotation in the PDTB-3 can be used as a basis for learning to classify implicit relations in languages that lack discourse annotation.
\citet{kim-etal-2020-implicit} explored whether the PDTB-3 could be used to learn fine-grained (Level-2) sense classification in general, while \citet{liang-etal-2020-extending} looked at whether separating inter-sentential implicits from intra-sentential implicits could improve their sense classification.
They also took a first step towards recognizing what sentences contained intra-sentential implicit relations, finding this benefitted from the use of linearized parse tree features.

Outside the PDTB-3 framework,  intra-sentential discourse relations are handled by (1) identifying discourse units (DUs), (2) attaching them to one another, and (3) associating the attachment with a coherence relation \cite{muller-etal-2012-constrained}.
One can therefore ask why we did not simply adopt this framework in the PDTB-3 and exploit the relatively good performance by systems in the DISRPT shared task on sentence-level discourse unit segmentation \cite{zeldes-etal-2019-disrpt}. There are two main reasons: First, DISRPT (and the approaches to discourse structure it covers) assumes that discourse segments cover a sentence with a non-overlapping partition. This is not the case with the PDTB, where the presence of overlapping segments (both within and across sentences) has been well documented \cite{lee2006complexity}. Second, discourse segments in these approaches are taken to correspond to syntactic units, which leads to both over-segmentation and under-segmentation in the PDTB-3. Of course, there are ``work-arounds'' for over-segmentation, such as RST's use of a \textsc{same-segment} relation \cite{mann1988rhetorical}, and under-segmentation can be addressed through additional segmentation. However, we decided that starting from scratch would allow us to clearly identify the problems of parsing intra-sentential implicits, at which point, we could consider what we could adopt from work done on the DISRPT shared task on sentence-level discourse unit segmentation \cite{zeldes-etal-2019-disrpt}.

\section{Methodology}

Given an input sentence $S$ represented as a sequence of tokens $s_1 \cdots s_n$, our aim is to identify the span of \argi[Arg1] and \argii[Arg2] if there exist an implicit discourse relation in that sentence and then to predict its corresponding sense relation. We treat the identification of argument spans as a sequence tagging problem, and the prediction of senses as a classification task. Thus, given $S$, our aim is to output both a tag sequence $Y$ of length $n$ and a sense label $c$ for the identified relation. The generated tag sequence $y_1 \cdots y_n$ contains token-level labels where $y_j \in \{$\texttt{B-Arg1}, \texttt{B-Arg2}, \texttt{I-Arg1}, \texttt{I-Arg2}, \texttt{O}$\}$ indicating whether the token belongs to \argi[Arg1], \argii[Arg2], or Other. We adopt the BIO format \citep{ramshaw1999text} since arguments (1) can span multiple tokens, (2) can occur in either order, (3) need not be adjacent, and (4) do not overlap. (Future work will address two additional properties of intra-sentential implicits: (1) as shown in Sec~\ref{sec:dataset}, a sentence can contain more than one such relation
and (2) even though most arguments are continuous spans, 264 intra-sentential implicit relations (4.2\%) have discontinuous spans.)

This section describes the two parts of our approach. Section~\ref{sec:dataset} describes the creation of two datasets based on the PDTB-3: $\mathcal{D}_1 = \{ (S^{(i)}, P^{(i)}, Y^{(i)}) \mid i \in \{ 1\ldots N \} \}$, $\mathcal{D}_2 = \{ (A_1^{(i)}, A_2^{(i)}, P^{(i)}, c^{(i)}) \mid i \in \{ 1\ldots M \} \}$ consisting of $N$ and $M$ input-output pairs, where $S$ is the input sentence, $P$ is its parse tree, $A_1$ and $A_2$ are \argi[Arg1] and \argii[Arg2] of the intra-sentential relation, $Y$ is the output sentence label, and $c$ is the sense label. Sections~\ref{sec:arg-ident}--\ref{sec:sense-class} describe models to recognize the argument spans and classify the relations. We provide detailed descriptions for these two steps in the rest of this section. 

\subsection{Dataset Generation}
\label{sec:dataset}

As we have two tasks, we built two datasets. Dataset $\mathcal{D}_1$ is used to train our argument identification models. It simply comprises individual sentences from the PDTB-3 and a sequence of labels of these sentences. Some models also contain parse tree features \cite{marcus-etal-1993-building}. To generate the sequence of labels $Y$, we take annotations of intra-sentential implicit relations from the PDTB-3. \argi[Arg1] tokens in the sentence are labelled \texttt{Arg1}, and \argii[Arg2] tokens are labelled \texttt{Arg2}. Tokens with \texttt{O} labels means they belong to neither \argi[Arg1] nor \argii[Arg2]. If a sentence does not have any intra-sentential implicit relation, then all of its tokens will be labelled \texttt{O}. In BIO format, these labels then become \texttt{B-Arg1}, \texttt{B-Arg2}, \texttt{I-Arg1}, \texttt{I-Arg2}, and \texttt{O}. 

\begin{table}[h]
\centering
\setlength\tabcolsep{4.5pt}
\begin{tabular}{crc}
Number of relations & Count & \% \\
\toprule
0 & 41,734 & 89.89\% \\
1 & 4,314 & 9.29\% \\
2 & 321 & 0.69\% \\
3 & 42 & 0.09\% \\
4 & 15 & 0.03\% \\
5 & 4 & 0.01\% \\
\bottomrule
total & 46,430 & 100\%
\end{tabular}
\caption{The distribution of intra-sentential implicit relations per sentence for our dataset.}
\label{tab:num_relations_distribution}
\end{table}

{
\begin{figure*}[t]
\centering
\includegraphics{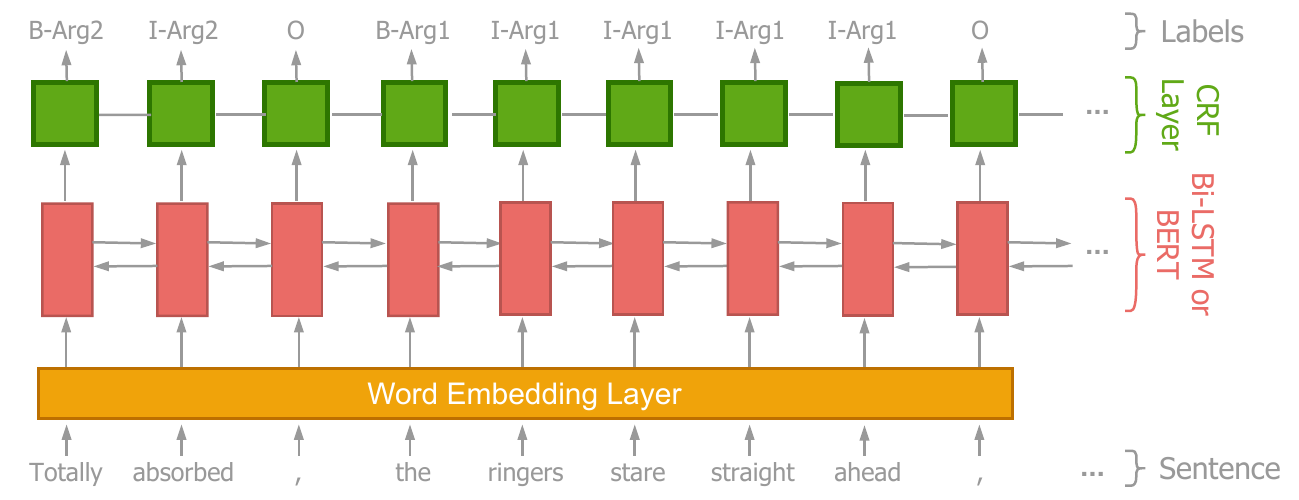}
\caption{\label{fig:model_architecture} Architecture of our proposed argument identification model.}
\end{figure*}
}

The dataset comprises the 46,430 sentences in the PDTB-3, with 24,369 intra-sentential relations, of which 6,234 are implicit.
Table~\ref{tab:num_relations_distribution} shows that a single sentence can have zero, one or more intra-sentential implicit relations, with over 99\% of the sentences having no more than one. So as not to lose any training data, the 321 sentences with two relations are duplicated, with each duplicate containing one of the relations. So while we do not currently try to learn multiple implicit relations within a sentence, this approach means we don't prejudice which relation is learned.

Another way of not losing data is to treat intra-sentential AltLex relations as intra-sentential implicits because they only differ from the latter in signalling its sense in its lexicalization \cite{liang-etal-2020-extending}. For instance, free-adjuncts are generally \argii[Arg2] of an intra-sentential implicit. However, those free adjuncts headed by ``avoiding'', ``contributing to'', ``resulting in'', etc. are labelled AltLex relations because the head uniquely signals a \textsc{Result} sense. Structurally, however, it is still an intra-sentential implicit.

Finally, the dataset does not include implicit relations that are linked to an explicit
relation. Such linking is used to convey that the arguments are semantically related in a way that cannot be attributed to the explicit discourse connective
alone \cite{pdtb3-manual:2019}.
A dedicated model to recognize implicits ``linked'' to explicit relations is included in work by
\citet{liang-etal-2020-extending}.


Dataset $\mathcal{D}_2$ is used to train the sense classifier. It only contains data on sentences with intra-sentential implicits, and is thus smaller than $\mathcal{D}_1$. Each entry in $\mathcal{D}_2$ includes \argi[Arg1], \argii[Arg2], the parse tree for the sentence in which they lie, and a sense label for
the \argi[Arg1]--\argii[Arg2] pair. Similar to $\mathcal{D}_1$, we also include AltLex relations.
The current effort uses Level-2 sense labels to avoid data sparsity, while still providing a more meaningful sense than the 4 coarse labels in Level-1. The distribution of Level-2 sense labels is shown in Table~\ref{tab:sense_label} in Appendix~\ref{appendix:dist_sense_label}.

\subsection{Argument Identification}
\label{sec:arg-ident}

The architecture for our argument identification model is shown in Figure~\ref{fig:model_architecture}. The input sentence first goes through the word embedding layer and then passes through either a BiLSTM \cite{hochreiter1997long} or BERT module. Then the learned representation over the input sentence is fed into a CRF layer to generate a sequence of labels $Y$. 

\paragraph{Baseline model}
The baseline model uses pre-trained GloVe \cite{pennington2014glove} vectors with BiLSTM and no additional parse tree features. For input sentence $S$ where $S = \{s_1, \ldots,s_n\}$ and $s_i$ denotes the $i$th token in $S$, the word vector $e_i$ is obtained from the word embedding module. Then a contextualized token-level encoding $h_i$ is obtained via a BiLSTM module:
{
\[ \overrightarrow{h}_i = \mathbf{LSTM}_f(e_i, \overrightarrow{h}_{i-1}),\]
\[ \overleftarrow{h}_i = \mathbf{LSTM}_b(e_i, \overleftarrow{h}_{i+1}),\]
\[ h_i = [\overrightarrow{h}_i;\overleftarrow{h}_i],\]
}
\noindent where $\overrightarrow{h}_i$ and $\overleftarrow{h}_i$ are hidden states of forward and backward LSTMs at time step $i$, and $;$ denotes concatenation.  

The resulting contextual word representations are then fed to the CRF layer to predict the $Y$ label.

\paragraph{BERT-based models}
\citet{shi-demberg-2019-next} observed that BERT-based models can benefit the task of classifying implicit \textit{inter-sentential} discourse relations. Here we ask whether they can help in the task of argument recognition for \textit{intra-sentential} implicit relations by creating variants of the baseline model where some parts of the model are replaced by BERT-based models. 

The first variant of the model we implemented replaces the pre-trained GloVe word vectors with a pre-trained BERT model for word embedding initialization. We then construct a second variant on top of this, which also contains parse tree features. (These come from Penn TreeBank parse trees \cite{marcus-etal-1993-building}, since we already know from \citet{liang-etal-2020-extending} that performance will drop when automated parse tree features are used.)  The parse trees are first linearized and then fed to a separate BiLSTM module. The learned parse tree representations are concatenated with learned representations of the input sentence to a single vector. This vector, containing both lexical and syntactic information of the input, is then fed to the CRF layer for output prediction.

These model variants use a pre-trained BERT model. We also implemented models that fine-tune BERT on our task. One variant uses the vanilla BERT model, replacing the BiLSTM module, and another variant has the same architecture but also uses parse tree features of the input sentences. 

{
\begin{figure}[t]
\centering
\includegraphics{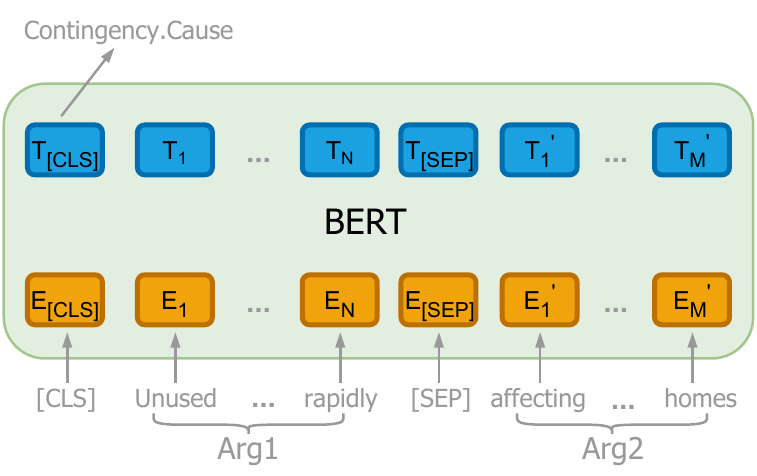}
\caption{\label{fig:sense_classifier} Architecture of our sense classifier.}
\end{figure}
}

\subsection{Sense Classification}
\label{sec:sense-class}

The sense classifier uses a BERT model whose input is the pair of arguments, \argi[Arg1] and \argii[Arg2], and whose output is a Level-2 discourse relation sense. The model architecture is illustrated in Figure~\ref{fig:sense_classifier}. Following \citet{shi-demberg-2019-next}, we concatenate \argi[Arg1] and \argii[Arg2] into a sequence separated with the special token [SEP]. We also insert the special token [CLS] at the beginning of the sequence. We use the output of BERT on the [CLS] token for sense classification. In addition, we also use parse trees as additional features to BERT. Similar to the method for argument identification, the learned parse tree features are concatenated with the representations of the [CLS] token to a single vector. This vector is then used for predicting the sense label.  

\subsection{Training and Inference}
Given the training set for argument identification with labelled sequence $\{ S^{(i)}, P^{(i)}, Y^{(i)} \mid i \in \{ 1\ldots N \} \}$, we maximize the conditional log likelihood for the sequence tagging objective:
\[ \bar{w} = \mathop{\mathrm{argmax}}_w \sum_{i=1}^{N} \log p(Y^{(i)} \mid S^{(i)}, P^{(i)}, w),\]
\noindent where $w$ denotes the model's parameters including the weights of the LSTM/BERT module and the transition weights of the CRF layer. The loss function for $Y$ labels is the negative log-likelihood based on $Y^{(i)} = \{y_1, \ldots, y_n\}$: 
\[ \mathcal{L}_\mathit{Y} = - \sum_{i=1}^{N} \sum_{j=1}^{n} \log p(y_j),\]

\noindent where $y_j\in Y^{(i)}$. 

For sense classification which is predicting $c$ label, the loss function is cross entropy:
\[ \mathcal{L}_\mathit{c} = - \sum_{i=1}^{M} \sum_{j=1}^{C} c^{(i)}_j \log p(c^{(i)}_j),\]
\noindent where $C$ denotes the number of classes.

At test time, inference for labels of a sentence $S$ involves applying the Viterbi algorithm to the CRF module to find the sequence with maximum likelihood $\hat{Y}$: 

\[ \hat{Y} = \mathop{\mathrm{argmax}}_Y P(Y \mid S, P, \bar{w}).\]

\section{Experiments}
For our experiments with LSTM-based models, we set the hidden dimensions to 256, the word embeddings to 100, and the vocabulary size to 50K. The word embeddings are initialized either using pre-trained Glove vectors (6B tokens, \textit{uncased}) or a pre-trained \textit{base-uncased} BERT. For experiments with BERT-based models, we use the same configuration with \textit{base-uncased} BERT from \citet{devlin-etal-2019-bert}. All of our training used the Adam optimizer \citep{KingmaB14}. LSTM-based models used a learning rate of 1e-3 and BERT-based models used a learning rate of 5e-5. We also use gradient clipping with a maximum gradient norm of 1 and we do not use any form of regularization. For sense classification, the gradient clipping is set with a maximum gradient norm of 0.5. 
We carry out assessment on the datasets mentioned in Section~\ref{sec:dataset} in two different ways: (1) using a random split of each dataset into training (60\%), development (20\%) and test (20\%) subsets, and (2) accepting the argument in \citet{shi-demberg-2017-need} that there is too much variation across the corpus for a single random split to produce representative results and that N-fold cross-validation will deliver more reliable and predictive results. In this work, we perform 10-fold cross-validation. 
We use loss on the development set to perform early stopping. All of our models were trained on a single Tesla P100 GPU with a batch size of 32. Our model implementation uses PyTorch \citep{NEURIPS2019_9015}. We used the BERT implementation from the Transformers library \citep{wolf-etal-2020-transformers}, and the CRF implementation from AllenNLP library \citep{DBLP:journals/corr/abs-1803-07640} for constrained decoding for the BIO scheme.

\begin{table*}[ht]
\centering
\begin{tabular}{|l|c|c|c|c|c|c|}
\hline
\multicolumn{1}{|c|}{\multirow{2}{*}{model}} & \multicolumn{3}{c|}{\argi[Arg1]} & \multicolumn{3}{c|}{\argii[Arg2]} \\ \cline{2-7} 
\multicolumn{1}{|c|}{}  & Precision      & Recall   & $F_{1}$         & Precision      & Recall   & $F_{1}$     \\ \hline
LSTM + GloVe (baseline)   & 35.50  & 30.73 & 32.79   & 43.14  & 41.57  & 42.06 \\ \hline
LSTM + BERT\textsuperscript{\dag}  & 39.83  & 49.01 & 43.44   & \textbf{56.93}  & 59.88  & \textbf{58.22} \\ \hline
LSTM + BERT\textsuperscript{\dag}  + parse tree  & 43.55  & 50.56 & 46.66    & 53.45 & \textbf{61.53} & 57.19 \\ \hline
BERT fine-tuned  & 45.80  & 50.63 & 48.06   & 54.56 & 59.04 & 56.64 \\ \hline
BERT fine-tuned  + parse tree & \textbf{49.62$\ast$}  & \textbf{51.39} & \textbf{50.43$\ast$}    & 56.47 & 58.31 & {57.28} \\ \hline
\end{tabular}
\caption{
Results of exact match for \argi[Arg1] and \argii[Arg2]. {\dag} denotes that a pre-trained BERT is used for word embedding, and $\ast$ denotes significant improvement over BERT fine-tuned with $p < 0.05$. Significance test was performed by estimating variance of the model from 5 random initialization.
}
\label{tab:all_model_exact_match}
\end{table*}

\begin{table*}[ht]
\small
\centering
\begin{tabular}{|l|c|c|c|c|c|c|}
\hline
\multicolumn{1}{|c|}{\multirow{2}{*}{model}} & \multicolumn{3}{c|}{\argi[Arg1]} & \multicolumn{3}{c|}{\argii[Arg2]} \\ \cline{2-7} 
\multicolumn{1}{|c|}{}                             & Precision      & Recall   & $F_{1}$         & Precision      & Recall   & $F_{1}$     \\ \hline
LSTM + BERT\textsuperscript{\dag} & 44.51$\pm$6.14  & 51.23$\pm$5.48 & 47.59$\pm$4.76    & 58.59$\pm$2.82  & 62.70$\pm$2.21 & 60.51$\pm$1.70 \\ \hline
LSTM + BERT\textsuperscript{\dag}  + parse tree  & 44.82$\pm$5.88  & 53.58$\pm$2.79 & 48.54$\pm$3.56    & \textbf{60.48$\pm$2.30}  & \textbf{61.64$\pm$2.68} & \textbf{61.00$\pm$1.70} \\ \hline
BERT fine-tuned & 48.22$\pm$5.55  & 53.26$\pm$2.94 & 50.42$\pm$3.58   & 58.59$\pm$3.29 & 58.64$\pm$2.24 & 58.64$\pm$1.86\\ \hline
BERT fine-tuned  + parse tree  & \textbf{49.74$\pm$3.41}  & \textbf{53.94$\pm$2.93} & \textbf{51.62$\pm$1.85}    & 60.17$\pm$2.49  & 60.37$\pm$2.52 & {60.22$\pm$1.78} \\ \hline
\end{tabular}
\caption{
Cross-validation results of exact match for \argi[Arg1] and \argii[Arg2]. {\dag} denotes that a pre-trained BERT is used for word embedding.
}
\label{tab:cross-valid_exact_match}
\end{table*}

\section{Results}
\paragraph{Argument Identification}
The best results here
on the sequence labelling objective for all model variants
come from the fine-tuned BERT-based model with additional parse tree features. 
In general, BERT-based models outperform LSTM-based models, and models that use parse tree features outperform those that don't.
 (All results appear in Appendix~\ref{appendix:argument_result}. The overall higher results  for \texttt{O} (i.e., neither \argi[Arg1] nor \argii[Arg2]) come simply from label imbalance.)

The standard automatic evaluation metrics for discourse argument recognition are Precision, Recall and $F_{1}$, computed on predicted arguments for relations that match the gold annotations. We follow \citet{xue-etal-2015-conll} in counting an argument as correctly recognized if and only if its span exactly matches the gold argument. No reward is given for partial match. Results on the test set from the 60/20/20 random split are shown for all models in Table~\ref{tab:all_model_exact_match}. Cross-validation results are shown in Table~\ref{tab:cross-valid_exact_match} for BERT-based models. The best overall performance comes from the  fine-tuned BERT model with parse tree features. 
The cross-validation results also show that the use of parse tree features improves Precision, if not Recall in all cases. This makes sense as the parse tree features can be used to reject what would otherwise be False Positives. In addition, we can see that for \argii[Arg2], the LSTM models with pre-trained BERT perform better than fine-tuned BERT. We haven't yet figured out any reason for this.
Finally, we can see that models perform better on \argii[Arg2] (both Recall and Precision) than on \argi[Arg1]. We attribute this to the fact that, even though \argii[Arg2] is not marked with an explicit connective (i.e., these are all implicit relations), there may still be positional and/or syntactic cues to its identity.

Supporting these observations are statistics from the test set. First, of its 987 sentences with intra-sentential implicit relations, the leftmost argument aligns with the beginning of the sentence 495 times (50.2\%). Of these, \argi[Arg1] is the leftmost argument 430 times (86.9\%). At the other end of the sentence, 685 (69.4\%) relations have their rightmost argument ending at the end of the sentence, almost 200 more than those with an argument at the beginning. Of these 685 relations, 614 (89.6\%) have \argii[Arg2] at their rightmost boundary. Note that whenever an argument starts or ends at a sentence boundary, the model just needs to predict the other end of the span, which is easier than predicting both ends. With \argii[Arg2] appearing more often at a sentence boundary than \argi[Arg1], this is consistent with our observation of model performance.


\begin{table*}[ht]
\centering
\begin{tabular}{|l|c|c|c|c|c|c|}
\hline
\multicolumn{1}{|c|}{\multirow{2}{*}{model}} & \multicolumn{3}{c|}{\argi[Arg1]-\argii[Arg2]} & \multicolumn{3}{c|}{\argii[Arg2]-\argi[Arg1]} \\ \cline{2-7} 
\multicolumn{1}{|c|}{}                              & Precision      & Recall  & $F_{1}$           & Precision      & Recall   & $F_{1}$    \\ \hline
LSTM + GloVe (baseline)              &  70.50  & 42.94  & 53.37  &  45.16  & 18.92  &  26.67 \\ \hline
LSTM + BERT\textsuperscript{\dag}                               & 70.03  & 75.03   & 74.01 & 51.85   & 56.76  & 54.19  \\ \hline
LSTM + BERT\textsuperscript{\dag} + parse tree                               & 70.82  & 70.97   & 70.90 & 39.22   & 27.03  & 32.00  \\ \hline
BERT fine-tuned                               & \textbf{83.77}  & 73.49   & \textbf{78.30} & 59.09   & 70.27  & 64.20  \\ \hline
BERT fine-tuned  + parse tree                           & 78.35  & \textbf{76.12} & 77.22    & \textbf{62.22}  & \textbf{75.68} & \textbf{68.29}  \\ \hline
\end{tabular}
\caption{
Results of argument order prediction with either \argi[Arg1]-\argii[Arg2] (914 instances from the test set) and \argii[Arg2]-\argi[Arg1] (73 instances from the test set). {\dag} denotes a pre-trained BERT is used for word embedding.
}
\label{tab:all_model_arg_order}
\end{table*}   

Even though the training data for the argument identification model contains at most one relation per sentence and the model only predicts continuous argument spans, we purposely relax the constraint for the model to predict at most one relation per sentence. We find that for all test set predictions, our best model predicts 77 sentences with more than one \argi[Arg1] (7.53\% of all sentences predicted to have a relation). It also predicts 44 sentences with more than one \argii[Arg2] (4.31\% of all sentences predicted to have a relation). This probably arises from duplicating those training instances with more than one relation (cf. Section~\ref{sec:dataset}). Later in this section, we show how the sense classifier can be used to choose among multiple predicted arguments.

\begin{table*}[ht]

\centering
\scalebox{1.0}{
\begin{tabular}{|l|ccc|ccc|c|}
\hline
\multicolumn{1}{|c|}{\multirow{2}{*}{condition}} & \multicolumn{3}{c|}{\argi[Arg1]} & \multicolumn{3}{c|}{\argii[Arg2]} &
\multicolumn{1}{|c|}{\multirow{2}{*}{\# of rels}}\\  
\multicolumn{1}{|c|}{}                             & Precision      & Recall   & $F_{1}$         & Precision      & Recall   & $F_{1}$    &  \multicolumn{1}{|c|}{} \\ \hline
 multiple relations                          & 31.25  & 28.57 & 29.85   & 21.95  & 25.71  & 23.68 & 36\\ 
 multiple relation + left                               & 25.00  & 23.53 & 24.24    & 10.00  & {11.76} & 10.81 & 18\\ 
 multiple relation + right                          & 37.50  & {33.33} & {35.29}    & 33.33  & 38.89 & {35.90} & 18\\ \hline
 Contingency.Cause                  & \textbf{68.12}  & \textbf{62.65} & \textbf{65.27}    & 75.44  & \textbf{69.08} & \textbf{72.12} & 249\\ 
 Contingency.Purpose             & 63.98  & 47.22 & 54.34  & \textbf{78.80}  & 57.54  & 66.51 & 252\\ 
 Expansion.Conjunction                   & 48.08  & 39.06 & 43.10    & 42.86  & 39.84 & 41.30 & 128\\ 
 Expansion.Level-of-detail               & 59.05  & 56.88 & 57.94    & 60.38  & 58.72 & 59.53 & 109\\ \hline
Original test set                   & 49.62  & {51.39} & {50.43}    & 56.47  & 58.31 & {57.28} & 987\\ \hline
\end{tabular}
}
\caption{
Results of argument identification for \argi[Arg1] and \argii[Arg2] for our best model under different conditions.  
}
\label{tab:best_model_analysis}
\end{table*}

As noted earlier, the order of arguments in an intra-sentential implicit relation is not fixed. So we have also correlated model performance with argument order.
Of the 5,157 sentences overall with intra-sentential implicit relations, 4,665 (90.5\%) have arguments in \argi[Arg1]-\argii[Arg2] order. In the 60/20/20 split  test set, this imbalance is higher across the 987 intra-sentential implicit relations, with 914 (92.6\%) showing the more common \argi[Arg1]-\argii[Arg2] order. Performance of the model in predicting argument order is shown in Table~\ref{tab:all_model_arg_order}. Note that a match of argument order here does not imply exact match of arguments in the previous evaluation. However, all models are more accurate in predicting the more frequent \argi[Arg1]-\argii[Arg2] order than \argii[Arg2]-\argi[Arg1] order,
although the performance difference is less with BERT-based models than with LSTM-based models, suggesting BERT can deal with under-represented data better.

\begin{table*}[t]
\centering
\scalebox{1.0}{
\setlength\tabcolsep{4.5pt}
\begin{tabular}{|l|ccc|}
\hline
\multicolumn{1}{|c|}{\multirow{2}{*}{Sense}} & \multicolumn{3}{c|}{BERT + Parse Tree} \\ 
\multicolumn{1}{|c|}{} & Precision & Recall & $F_{1}$ \\\hline
Contingency.Cause   &  76.50 $\pm$ 4.22 &	74.60 $\pm$ 4.89 &	75.31 $\pm$ 2.14           \\
Contingency.Purpose   &   86.78 $\pm$ 2.05 &	90.49 $\pm$	2.66 &	88.56 $\pm$	1.57        \\
Expansion.Conjunction  & 73.99 $\pm$ 8.05 &	70.33 $\pm$	5.55 &	71.82 $\pm$	5.03            \\
Expansion.Level-of-detail  &  57.79	$\pm$ 4.70 &	48.23 $\pm$	3.12 &	52.37 $\pm$	2.06     \\
\hline
Overall & 70.86 $\pm$	1.45 &	69.54 $\pm$	1.30 &	69.65 $\pm$	1.36 \\
\hline
\end{tabular}
}
\caption{Cross-validation results for sense classification.} 
\label{tab:sense_classification_cross_validation}
\end{table*}



For completeness, we also analyze the best model's argument identification performance under different conditions. First, we consider that part of the 60/20/20 split test set whose sentences contain more than one intra-sentential implicit relations. For a fair comparison, we ensure that all single-relation tokens derived from the same sentence are in the test set (36 tokens). We further divide these tokens into ones whose relation is further left in the sentence and ones whose relation is further right. The top part of Table~\ref{tab:best_model_analysis} shows the performance of the fine-tuned BERT model with parse tree features on these sets. Compared with the original test set, the overall performance on relations derived from sentences with multiple relations is much worse. This is expected as the training set for argument identification assumed that a sentence can only have at most one intra-sentential implicit relation.

We next examine performance for different sense labels, based on the four most frequent senses in the test set -- those that appear more than 100 times. The bottom part of Table~\ref{tab:best_model_analysis} shows that our model performed best on \textsc{Contingency.Cause}. Inspection of the True Positives shows this to result from \argii[Arg2] of these relations often being headed by a AltLex token with Part-of-speech tag VBG, which uniquely conveys the sense \textsc{Contingency.Cause.Result}. The model can thus easily learn to recognize these arguments using the parse tree features. 





\paragraph{Sense Classification} Cross-validation results for Precision, Recall, and $F_1$ for the top four senses recognized by our sense classifier are shown in Table~\ref{tab:sense_classification_cross_validation}. The complete performance breakdown is given in Table~\ref{tab:sense_classification_result_full_cross_validation} in Appendix~\ref{appendix:sense_result}, along with a confusion matrix (Figure~\ref{fig:confusion_matrix}) and results on the test set in the 60/20/20 random split (Table~\ref{tab:sense_classification_result_full}). As the distribution of senses is imbalanced, we calculate overall performance by averaging performance for each sense, weighted by its frequency in the test set.
Note that the overall $F_1$ for intra-sentential implicits is much higher than 50.41 reported in \citet{liang-etal-2020-extending} where a LSTM-based model is used.
The overall accuracy of our sense classifier on cross-validation is 69.54\%, and that on the test set in the 60/20/20 random split is 75.19\%. From the results, we can see that the performance for \textsc{Contingency.Purpose} is very high, with $F_1$ exceeding the weighted average by almost 20 points.
We speculate that this is due to the fact that over 90\% of \textsc{Contingency.Purpose} labels are on relations where \argii[Arg2] begins with a free ``to clause''
and there are few other intra-sentential implicits labelled \textsc{Contingency.Purpose}.

Our sense classifier is trained using gold argument spans taken from the PDTB-3. In reality, such gold annotations will rarely be available. Thus, we also use predicted arguments from our argument identification model as inputs to test the ability of our sense classifier. Specifically, we first obtain test set sentences with intra-sentential implicit relations and their parse trees. Then we feed those to our argument identification model to get predicted arguments. These predicted arguments are then fed in to our sense classifier together with the parse tree features. If the argument identification model fails to predict any arguments for any sentences, we ignore them in our evaluation. Table~\ref{tab:sense_classification_result_pred_vs_gold_full} in Appendix~\ref{appendix:sense_result} shows the sense classification results using predicted arguments and, for comparison, the results using the gold arguments. Note that as we dropped some sentences, the results with gold arguments are slightly different from those of the original test set. 
Because the performance drop is small going from gold arguments to predicted arguments, we can argue that our models can be used as a pipeline for handling intra-sentential implicit relations in shallow discourse parsing with the input simply being a sentence.  

We noted in Section~\ref{sec:dataset} that our argument identification model might predict multiple \argi[Arg1]s and/or \argii[Arg2]s for a given sentence. We therefore assessed whether the sense classifier could be used to decide which of the predicted arguments to use. Specifically, we identify cases where one argument has a single prediction but there are multiple predictions for the other argument. For each \argi[Arg1]-\argii[Arg2] pair, we use the sense classifier to predict the sense label and its likelihood. Comparing these likelihoods, we select the pair with the highest certainty, ignoring cases where the predicted senses are the same for all pairs. 
We also implement a baseline of always choosing the pair with the most frequent sense, which is \textsc{Contingency.Cause}. The 60/20/20 split test set shows 34 cases in which there are multiple predictions of \argi[Arg1] for a single \argii[Arg2]. In 23 out of these 34 instances, the correct \argi[Arg1] is associated with the pair with the highest likelihood. The baseline only gets 13 cases correct. Similarly, the test set shows 23 cases in which there are multiple predictions of \argii[Arg2] for a single \argi[Arg1]. In 13 of these 23 instances, the pair for which the sense classifier assigns the highest priority correctly identifies which \argii[Arg2] to use. The baseline only gets 9 cases correct. While further analysis is needed, this does show that the sense classifier can contribute to selecting the right argument from the set of predicted argument candidates.


\section{Conclusions and Future Work}
\label{sec:concl}

To the best of our knowledge, this is the first work to attempt to identify the arguments of intra-sentential implicit discourse relations in the framework of the PDTB-3, as well as their order and at least one of their sense relations. We used a model architecture similar to models of a sequence tagging task, and concluded that BERT-based models have better performance than LSTM-based models in both exactly matching gold annotations of arguments and correctly predicting the order of \argi[Arg1] and \argii[Arg2]. We confirmed that using parse trees features as input to the model assists with these tasks. We also provide evidence that our sense classifier, together with the argument recognizer, can be used as a pipeline for handling intra-sentential implicit relations. We also find that the sense classifier can be used to aid the selection of the right argument from the set of predicted argument candidates.

Our methods have several limitations. First, we assumed every sentence can have at most one intra-sentential implicit relation, whereas in reality, multiple such relations are possible (cf. Table~\ref{tab:num_relations_distribution}). Secondly, our approach does not support the case of discontinuous argument spans. Thirdly, we have ignored implicit relations ``linked'' to explicit relations (either intra-sentential or inter-sentential).
Finally, although we have followed the lead of \citet{shi-demberg-2017-need} in assessing performance using cross-validation because it is more reliable than choosing a specific test set, we did not cast all our results in those terms. In the future, we plan to address these problems and develop methods that can identify all existing relations within the sentence.

\section*{Acknowledgments}
This work was supported in part by the UKRI Centre for Doctoral Training in Natural Language Processing (UKRI grant EP/S022481/1), the University of Edinburgh, and Huawei. We would like to thank Hannah Rohde and anonymous CODI reviewers for their helpful feedback.

\bibliographystyle{acl_natbib}
\bibliography{anthology,custom}

\begin{thebibliography}{37}
\expandafter\ifx\csname natexlab\endcsname\relax\def\natexlab#1{#1}\fi

\bibitem[{Bai and Zhao(2018)}]{bai-zhao-2018-deep}
Hongxiao Bai and Hai Zhao. 2018.
\newblock \href {https://www.aclweb.org/anthology/C18-1048} {Deep enhanced
  representation for implicit discourse relation recognition}.
\newblock In \emph{Proceedings of the 27th International Conference on
  Computational Linguistics}, pages 571--583, Santa Fe, New Mexico, USA.
  Association for Computational Linguistics.

\bibitem[{Cohan et~al.(2018)Cohan, Dernoncourt, Kim, Bui, Kim, Chang, and
  Goharian}]{cohan-etal-2018-discourse}
Arman Cohan, Franck Dernoncourt, Doo~Soon Kim, Trung Bui, Seokhwan Kim, Walter
  Chang, and Nazli Goharian. 2018.
\newblock \href {https://doi.org/10.18653/v1/N18-2097} {A discourse-aware
  attention model for abstractive summarization of long documents}.
\newblock In \emph{Proceedings of the 2018 Conference of the North {A}merican
  Chapter of the Association for Computational Linguistics: Human Language
  Technologies, Volume 2 (Short Papers)}, pages 615--621, New Orleans,
  Louisiana. Association for Computational Linguistics.

\bibitem[{Dai and Huang(2018)}]{dai-huang-2018-improving}
Zeyu Dai and Ruihong Huang. 2018.
\newblock \href {https://doi.org/10.18653/v1/N18-1013} {Improving implicit
  discourse relation classification by modeling inter-dependencies of discourse
  units in a paragraph}.
\newblock In \emph{Proceedings of the 2018 Conference of the North {A}merican
  Chapter of the Association for Computational Linguistics: Human Language
  Technologies, Volume 1 (Long Papers)}, pages 141--151, New Orleans,
  Louisiana. Association for Computational Linguistics.

\bibitem[{Devlin et~al.(2019)Devlin, Chang, Lee, and
  Toutanova}]{devlin-etal-2019-bert}
Jacob Devlin, Ming-Wei Chang, Kenton Lee, and Kristina Toutanova. 2019.
\newblock \href {https://doi.org/10.18653/v1/N19-1423} {{BERT}: Pre-training of
  deep bidirectional transformers for language understanding}.
\newblock In \emph{Proceedings of the 2019 Conference of the North {A}merican
  Chapter of the Association for Computational Linguistics: Human Language
  Technologies, Volume 1 (Long and Short Papers)}, pages 4171--4186,
  Minneapolis, Minnesota. Association for Computational Linguistics.

\bibitem[{Gardner et~al.(2018)Gardner, Grus, Neumann, Tafjord, Dasigi, Liu,
  Peters, Schmitz, and Zettlemoyer}]{DBLP:journals/corr/abs-1803-07640}
Matt Gardner, Joel Grus, Mark Neumann, Oyvind Tafjord, Pradeep Dasigi,
  Nelson~F. Liu, Matthew~E. Peters, Michael Schmitz, and Luke Zettlemoyer.
  2018.
\newblock \href {http://arxiv.org/abs/1803.07640} {Allennlp: {A} deep semantic
  natural language processing platform}.
\newblock \emph{CoRR}, abs/1803.07640.

\bibitem[{Hochreiter and Schmidhuber(1997)}]{hochreiter1997long}
Sepp Hochreiter and J\"{u}rgen Schmidhuber. 1997.
\newblock \href {https://doi.org/10.1162/neco.1997.9.8.1735} {Long short-term
  memory}.
\newblock \emph{Neural Comput.}, 9(8):1735–1780.

\bibitem[{Jansen et~al.(2014)Jansen, Surdeanu, and
  Clark}]{jansen-etal-2014-discourse}
Peter Jansen, Mihai Surdeanu, and Peter Clark. 2014.
\newblock \href {https://doi.org/10.3115/v1/P14-1092} {Discourse complements
  lexical semantics for non-factoid answer reranking}.
\newblock In \emph{Proceedings of the 52nd Annual Meeting of the Association
  for Computational Linguistics (Volume 1: Long Papers)}, pages 977--986,
  Baltimore, Maryland. Association for Computational Linguistics.

\bibitem[{Joty et~al.(2012)Joty, Carenini, and Ng}]{joty-etal-2012-novel}
Shafiq Joty, Giuseppe Carenini, and Raymond Ng. 2012.
\newblock \href {https://www.aclweb.org/anthology/D12-1083} {A novel
  discriminative framework for sentence-level discourse analysis}.
\newblock In \emph{Proceedings of the 2012 Joint Conference on Empirical
  Methods in Natural Language Processing and Computational Natural Language
  Learning}, pages 904--915, Jeju Island, Korea. Association for Computational
  Linguistics.

\bibitem[{Kim et~al.(2020)Kim, Feng, Gunasekara, and
  Lastras}]{kim-etal-2020-implicit}
Najoung Kim, Song Feng, Chulaka Gunasekara, and Luis Lastras. 2020.
\newblock \href {https://doi.org/10.18653/v1/2020.acl-main.480} {Implicit
  discourse relation classification: We need to talk about evaluation}.
\newblock In \emph{Proceedings of the 58th Annual Meeting of the Association
  for Computational Linguistics}, pages 5404--5414, Online. Association for
  Computational Linguistics.

\bibitem[{Kingma and Ba(2015)}]{KingmaB14}
Diederik~P. Kingma and Jimmy Ba. 2015.
\newblock \href {http://arxiv.org/abs/1412.6980} {Adam: {A} method for
  stochastic optimization}.
\newblock In \emph{3rd International Conference on Learning Representations,
  {ICLR} 2015, San Diego, CA, USA, May 7-9, 2015, Conference Track
  Proceedings}.

\bibitem[{Kurfal{\i} and {\"O}stling(2019)}]{kurfali-ostling-2019-zero}
Murathan Kurfal{\i} and Robert {\"O}stling. 2019.
\newblock \href {https://doi.org/10.18653/v1/W19-5927} {Zero-shot transfer for
  implicit discourse relation classification}.
\newblock In \emph{Proceedings of the 20th Annual SIGdial Meeting on Discourse
  and Dialogue}, pages 226--231, Stockholm, Sweden. Association for
  Computational Linguistics.

\bibitem[{Lafferty et~al.(2001)Lafferty, McCallum, and Pereira}]{LaffertyMP01}
John~D. Lafferty, Andrew McCallum, and Fernando C.~N. Pereira. 2001.
\newblock Conditional random fields: Probabilistic models for segmenting and
  labeling sequence data.
\newblock In \emph{Proceedings of the Eighteenth International Conference on
  Machine Learning {(ICML} 2001), Williams College, Williamstown, MA, USA, June
  28 - July 1, 2001}, pages 282--289. Morgan Kaufmann.

\bibitem[{Lee et~al.(2006)Lee, Prasad, Joshi, Dinesh, and
  Webber}]{lee2006complexity}
Alan Lee, Rashmi Prasad, Aravind Joshi, Nikhil Dinesh, and Bonnie Webber. 2006.
\newblock Complexity of dependencies in discourse: Are dependencies in
  discourse more complex than in syntax.
\newblock In \emph{Proceedings of the 5th International Workshop on Treebanks
  and Linguistic Theories}, pages 12--23.

\bibitem[{Li et~al.(2014)Li, Carpuat, and Nenkova}]{li-etal-2014-assessing}
Junyi~Jessy Li, Marine Carpuat, and Ani Nenkova. 2014.
\newblock \href {https://doi.org/10.3115/v1/P14-2047} {Assessing the discourse
  factors that influence the quality of machine translation}.
\newblock In \emph{Proceedings of the 52nd Annual Meeting of the Association
  for Computational Linguistics (Volume 2: Short Papers)}, pages 283--288,
  Baltimore, Maryland. Association for Computational Linguistics.

\bibitem[{Liang et~al.(2020)Liang, Zhao, and
  Webber}]{liang-etal-2020-extending}
Li~Liang, Zheng Zhao, and Bonnie Webber. 2020.
\newblock \href {https://doi.org/10.18653/v1/2020.codi-1.14} {Extending
  implicit discourse relation recognition to the {PDTB}-3}.
\newblock In \emph{Proceedings of the First Workshop on Computational
  Approaches to Discourse}, pages 135--147, Online. Association for
  Computational Linguistics.

\bibitem[{Lin et~al.(2014)Lin, Ng, and Kan}]{lin2014pdtb}
Ziheng Lin, Hwee~Tou Ng, and Min-Yen Kan. 2014.
\newblock \href {https://doi.org/10.1017/S1351324912000307} {A pdtb-styled
  end-to-end discourse parser}.
\newblock \emph{Natural Language Engineering}, 20(2):151--184.

\bibitem[{Liu and Li(2016)}]{liu-li-2016-recognizing}
Yang Liu and Sujian Li. 2016.
\newblock \href {https://doi.org/10.18653/v1/D16-1130} {Recognizing implicit
  discourse relations via repeated reading: Neural networks with multi-level
  attention}.
\newblock In \emph{Proceedings of the 2016 Conference on Empirical Methods in
  Natural Language Processing}, pages 1224--1233, Austin, Texas. Association
  for Computational Linguistics.

\bibitem[{Mann and Thompson(1988)}]{mann1988rhetorical}
William~C Mann and Sandra~A Thompson. 1988.
\newblock Rhetorical structure theory: Toward a functional theory of text
  organization.
\newblock \emph{Text}, 8(3):243--281.

\bibitem[{Marcus et~al.(1993)Marcus, Santorini, and
  Marcinkiewicz}]{marcus-etal-1993-building}
Mitchell~P. Marcus, Beatrice Santorini, and Mary~Ann Marcinkiewicz. 1993.
\newblock \href {https://www.aclweb.org/anthology/J93-2004} {Building a large
  annotated corpus of {E}nglish: The {P}enn {T}reebank}.
\newblock \emph{Computational Linguistics}, 19(2):313--330.

\bibitem[{Muller et~al.(2012)Muller, Afantenos, Denis, and
  Asher}]{muller-etal-2012-constrained}
Philippe Muller, Stergos Afantenos, Pascal Denis, and Nicholas Asher. 2012.
\newblock \href {https://www.aclweb.org/anthology/C12-1115} {Constrained
  decoding for text-level discourse parsing}.
\newblock In \emph{Proceedings of {COLING} 2012}, pages 1883--1900, Mumbai,
  India. The COLING 2012 Organizing Committee.

\bibitem[{Paszke et~al.(2019)Paszke, Gross, Massa, Lerer, Bradbury, Chanan,
  Killeen, Lin, Gimelshein, Antiga, Desmaison, Kopf, Yang, DeVito, Raison,
  Tejani, Chilamkurthy, Steiner, Fang, Bai, and Chintala}]{NEURIPS2019_9015}
Adam Paszke, Sam Gross, Francisco Massa, Adam Lerer, James Bradbury, Gregory
  Chanan, Trevor Killeen, Zeming Lin, Natalia Gimelshein, Luca Antiga, Alban
  Desmaison, Andreas Kopf, Edward Yang, Zachary DeVito, Martin Raison, Alykhan
  Tejani, Sasank Chilamkurthy, Benoit Steiner, Lu~Fang, Junjie Bai, and Soumith
  Chintala. 2019.
\newblock \href
  {http://papers.neurips.cc/paper/9015-pytorch-an-imperative-style-high-performance-deep-learning-library.pdf}
  {Pytorch: An imperative style, high-performance deep learning library}.
\newblock In H.~Wallach, H.~Larochelle, A.~Beygelzimer, F.~d~Alch\'{e}-Buc,
  E.~Fox, and R.~Garnett, editors, \emph{Advances in Neural Information
  Processing Systems 32}, pages 8024--8035. Curran Associates, Inc.

\bibitem[{Pennington et~al.(2014)Pennington, Socher, and
  Manning}]{pennington2014glove}
Jeffrey Pennington, Richard Socher, and Christopher~D. Manning. 2014.
\newblock \href {http://www.aclweb.org/anthology/D14-1162} {Glove: Global
  vectors for word representation}.
\newblock In \emph{Empirical Methods in Natural Language Processing (EMNLP)},
  pages 1532--1543.

\bibitem[{Prasad et~al.(2008)Prasad, Dinesh, Lee, Miltsakaki, Robaldo, Joshi,
  and Webber}]{prasad-etal-2008-penn}
Rashmi Prasad, Nikhil Dinesh, Alan Lee, Eleni Miltsakaki, Livio Robaldo,
  Aravind Joshi, and Bonnie Webber. 2008.
\newblock \href
  {http://www.lrec-conf.org/proceedings/lrec2008/pdf/754_paper.pdf} {The {P}enn
  {D}iscourse {T}ree{B}ank 2.0.}
\newblock In \emph{Proceedings of the Sixth International Conference on
  Language Resources and Evaluation ({LREC}'08)}, Marrakech, Morocco. European
  Language Resources Association (ELRA).

\bibitem[{Prasad et~al.(2017)Prasad, Forbes~Riley, and
  Lee}]{prasad-etal-2017-towards}
Rashmi Prasad, Katherine Forbes~Riley, and Alan Lee. 2017.
\newblock \href {https://doi.org/10.18653/v1/W17-5502} {Towards full text
  shallow discourse relation annotation: Experiments with cross-paragraph
  implicit relations in the {PDTB}}.
\newblock In \emph{Proceedings of the 18th Annual {SIG}dial Meeting on
  Discourse and Dialogue}, pages 7--16, Saarbr{\"u}cken, Germany. Association
  for Computational Linguistics.

\bibitem[{Prasad et~al.(2010)Prasad, Joshi, and
  Webber}]{prasad-etal-2010-realization}
Rashmi Prasad, Aravind Joshi, and Bonnie Webber. 2010.
\newblock \href {https://www.aclweb.org/anthology/C10-2118} {Realization of
  discourse relations by other means: Alternative lexicalizations}.
\newblock In \emph{Coling 2010: Posters}, pages 1023--1031, Beijing, China.
  Coling 2010 Organizing Committee.

\bibitem[{Ramshaw and Marcus(1999)}]{ramshaw1999text}
Lance~A Ramshaw and Mitchell~P Marcus. 1999.
\newblock Text chunking using transformation-based learning.
\newblock In \emph{Natural language processing using very large corpora}, pages
  157--176. Springer.

\bibitem[{Shi and Demberg(2017)}]{shi-demberg-2017-need}
Wei Shi and Vera Demberg. 2017.
\newblock \href {https://www.aclweb.org/anthology/E17-2024} {On the need of
  cross validation for discourse relation classification}.
\newblock In \emph{Proceedings of the 15th Conference of the {E}uropean Chapter
  of the Association for Computational Linguistics: Volume 2, Short Papers},
  pages 150--156, Valencia, Spain. Association for Computational Linguistics.

\bibitem[{Shi and Demberg(2019)}]{shi-demberg-2019-next}
Wei Shi and Vera Demberg. 2019.
\newblock \href {https://doi.org/10.18653/v1/D19-1586} {Next sentence
  prediction helps implicit discourse relation classification within and across
  domains}.
\newblock In \emph{Proceedings of the 2019 Conference on Empirical Methods in
  Natural Language Processing and the 9th International Joint Conference on
  Natural Language Processing (EMNLP-IJCNLP)}, pages 5790--5796, Hong Kong,
  China. Association for Computational Linguistics.

\bibitem[{Stede(2011)}]{stede2011discourse}
Manfred Stede. 2011.
\newblock Discourse processing.
\newblock \emph{Synthesis Lectures on Human Language Technologies},
  4(3):1--165.

\bibitem[{Stede et~al.(2016)Stede, Afantenos, Peldszus, Asher, and
  Perret}]{stede-etal-2016-parallel}
Manfred Stede, Stergos Afantenos, Andreas Peldszus, Nicholas Asher, and
  J{\'e}r{\'e}my Perret. 2016.
\newblock \href {https://www.aclweb.org/anthology/L16-1167} {Parallel discourse
  annotations on a corpus of short texts}.
\newblock In \emph{Proceedings of the Tenth International Conference on
  Language Resources and Evaluation ({LREC}'16)}, pages 1051--1058,
  Portoro{\v{z}}, Slovenia. European Language Resources Association (ELRA).

\bibitem[{Varia et~al.(2019)Varia, Hidey, and
  Chakrabarty}]{varia-etal-2019-discourse}
Siddharth Varia, Christopher Hidey, and Tuhin Chakrabarty. 2019.
\newblock \href {https://doi.org/10.18653/v1/W19-5951} {Discourse relation
  prediction: Revisiting word pairs with convolutional networks}.
\newblock In \emph{Proceedings of the 20th Annual SIGdial Meeting on Discourse
  and Dialogue}, pages 442--452, Stockholm, Sweden. Association for
  Computational Linguistics.

\bibitem[{Wang and Lan(2015)}]{wang-lan-2015-refined}
Jianxiang Wang and Man Lan. 2015.
\newblock \href {https://doi.org/10.18653/v1/K15-2002} {A refined end-to-end
  discourse parser}.
\newblock In \emph{Proceedings of the Nineteenth Conference on Computational
  Natural Language Learning - Shared Task}, pages 17--24, Beijing, China.
  Association for Computational Linguistics.

\bibitem[{Webber et~al.(2019)Webber, Prasad, Lee, and
  Joshi}]{pdtb3-manual:2019}
Bonnie Webber, Rashmi Prasad, Alan Lee, and Aravind Joshi. 2019.
\newblock The penn discourse treebank 3.0 annotation manual.
\newblock
  \url{https://catalog.ldc.upenn.edu/docs/LDC2019T05/PDTB3-Annotation-Manual.pdf}.

\bibitem[{Wolf et~al.(2020)Wolf, Debut, Sanh, Chaumond, Delangue, Moi, Cistac,
  Rault, Louf, Funtowicz, Davison, Shleifer, von Platen, Ma, Jernite, Plu, Xu,
  Le~Scao, Gugger, Drame, Lhoest, and Rush}]{wolf-etal-2020-transformers}
Thomas Wolf, Lysandre Debut, Victor Sanh, Julien Chaumond, Clement Delangue,
  Anthony Moi, Pierric Cistac, Tim Rault, Remi Louf, Morgan Funtowicz, Joe
  Davison, Sam Shleifer, Patrick von Platen, Clara Ma, Yacine Jernite, Julien
  Plu, Canwen Xu, Teven Le~Scao, Sylvain Gugger, Mariama Drame, Quentin Lhoest,
  and Alexander Rush. 2020.
\newblock \href {https://doi.org/10.18653/v1/2020.emnlp-demos.6} {Transformers:
  State-of-the-art natural language processing}.
\newblock In \emph{Proceedings of the 2020 Conference on Empirical Methods in
  Natural Language Processing: System Demonstrations}, pages 38--45, Online.
  Association for Computational Linguistics.

\bibitem[{Xue et~al.(2015)Xue, Ng, Pradhan, Prasad, Bryant, and
  Rutherford}]{xue-etal-2015-conll}
Nianwen Xue, Hwee~Tou Ng, Sameer Pradhan, Rashmi Prasad, Christopher Bryant,
  and Attapol Rutherford. 2015.
\newblock \href {https://doi.org/10.18653/v1/K15-2001} {The {C}o{NLL}-2015
  shared task on shallow discourse parsing}.
\newblock In \emph{Proceedings of the Nineteenth Conference on Computational
  Natural Language Learning - Shared Task}, pages 1--16, Beijing, China.
  Association for Computational Linguistics.

\bibitem[{Xue et~al.(2016)Xue, Ng, Pradhan, Rutherford, Webber, Wang, and
  Wang}]{xue-etal-2016-conll}
Nianwen Xue, Hwee~Tou Ng, Sameer Pradhan, Attapol Rutherford, Bonnie Webber,
  Chuan Wang, and Hongmin Wang. 2016.
\newblock \href {https://doi.org/10.18653/v1/K16-2001} {{C}o{NLL} 2016 shared
  task on multilingual shallow discourse parsing}.
\newblock In \emph{Proceedings of the {C}o{NLL}-16 shared task}, pages 1--19,
  Berlin, Germany. Association for Computational Linguistics.

\bibitem[{Zeldes et~al.(2019)Zeldes, Das, Maziero, Antonio, and
  Iruskieta}]{zeldes-etal-2019-disrpt}
Amir Zeldes, Debopam Das, Erick~Galani Maziero, Juliano Antonio, and Mikel
  Iruskieta. 2019.
\newblock \href {https://doi.org/10.18653/v1/W19-2713} {The {DISRPT} 2019
  shared task on elementary discourse unit segmentation and connective
  detection}.
\newblock In \emph{Proceedings of the Workshop on Discourse Relation Parsing
  and Treebanking 2019}, pages 97--104, Minneapolis, MN. Association for
  Computational Linguistics.

\end{thebibliography}

\clearpage
\pagebreak
\newpage

\appendix

\section{Appendix}
\subsection{Distribution of Sense Label}
\label{appendix:dist_sense_label}
The distribution of intra-sentential implicit sense labels is shown in Table~\ref{tab:sense_label}. We can see that the sense labels are very imbalanced.

\begin{table}[ht]
\centering
\setlength\tabcolsep{4.5pt}
\begin{tabular}{|lrr|}
\hline
Sense & Count & \% \\
\hline
Comparison.Concession      &          66 & 1.28\% \\
Comp.Concession+S.A.  &     4 & 0.08\% \\
Comparison.Contrast      &           112 & 2.17\% \\
Comparison.Similarity      &           7 & 0.14\% \\
\hline
Contingency.Cause   &               1366 & 26.49\%\\
Contingency.Cause+Belief   &          66 & 1.28\% \\
Cont.Cause+SpeechAct       &    1 & 0.02\% \\
Contingency.Condition    &           222 & 4.30\% \\
Cont.Condition+SpeechAct   &    1 & 0.02\% \\
Cont.Negative-condition     &   1 & 0.02\% \\
Contingency.Purpose   &             1323 & 25.65\% \\
\hline
Expansion.Conjunction  &             667 & 12.93\% \\
Expansion.Disjunction       &         17 & 0.33\% \\
Expansion.Equivalence      &          35 & 0.68\% \\
Expansion.Instantiation   &           86 & 1.67\% \\
Expansion.Level-of-detail  &         565 & 10.96\% \\
Expansion.Manner         &           183 & 3.55\% \\
Expansion.Substitution    &           82 & 1.59\% \\
\hline
Temporal.Asynchronous    &           178 & 3.45\% \\
Temporal.Synchronous     &           175 & 3.39\% \\
\hline
\end{tabular}
\caption{The distribution of Level-2 labels for intra-sentential implicit relations in our dataset. Comp is short for Comparison, Cont is short for Contingency, and S.A. is short for SpeechAct.}
\label{tab:sense_label}
\end{table}

\subsection{Additional Results for Argument Identification}
\label{appendix:argument_result}
In this section, we provide the results of the sequence tagging objective for all models variants as well as the results of argument order prediction. All results are reported on the test set. Table~\ref{tab:lstm_glove} shows results for our baseline LSTM model using GloVe word embeddings. Table~\ref{tab:lstm_bert} then shows results for the model when GloVe word embedding are replacrd with BERT embeddings. There is a large increase in Recall, with no drop in Precision except for \texttt{O} labels. Table~\ref{tab:lstm_bert_tree} shows results when parse tree features are added to the model. Here, there is an increase in Recall for \argi[Arg1], and while Recall for \argii[Arg2] decreases somewhat, it is accompanied by an increase in Precision. Then, we experiment with the vanilla BERT model and fine-tuning it on our dataset. The results are provided in Table~\ref{tab:bert_finetune}. We can again observe a slight increase in performance, showcasing the power of BERT, even without the parse tree features. Finally, the last model we implement is BERT taking additional parse tree features fine-tuned on our dataset. The results are shown in Table~\ref{tab:bert_finetune_tree}. We can see that this is the best performing model for argument identification, proving that both BERT and parse tree features can improve performance. 

\begin{table}[ht]
\centering
\begin{tabular}{|c|c|c|c|}
\hline
{Label} & {Precision} & {Recall} & {$F_1$}\\
\hline
\texttt{B-Arg1} & 46.50 & 35.62 & 40.43\\
\texttt{B-Arg2} & 54.07 & 46.76 & 50.15\\
\texttt{I-Arg1} & 54.46 & 46.51 & 50.17\\
\texttt{I-Arg2} & 48.64 & 60.03 & 53.74\\
\texttt{O} & 95.38 & 95.24 & 95.31\\
\hline
\end{tabular}
\caption{
Results of the baseline LSTM model on test set using GloVe word embedding. 
}
\label{tab:lstm_glove}
\end{table}

\begin{table}[ht]
\centering
\begin{tabular}{|c|c|c|c|}
\hline
{Label} & {Precision} & {Recall} & {$F_1$}\\
\hline
\texttt{B-Arg1} & 48.10 & 52.35 & 50.14\\
\texttt{B-Arg2} & 60.88 & 69.11 & 64.73\\
\texttt{I-Arg1} & 63.34 & 65.59 & 64.45\\
\texttt{I-Arg2} & 67.42 & 70.53 & 68.94\\
\texttt{O} & 88.84 & 96.47 & 92.50\\
\hline
\end{tabular}
\caption{
Results of the LSTM model on test set using BERT word embedding. 
}
\label{tab:lstm_bert}
\end{table}

\begin{table}[ht]
\centering
\begin{tabular}{|c|c|c|c|}
\hline
{Label} & {Precision} & {Recall} & {$F_1$}\\
\hline
\texttt{B-Arg1} & 50.29 & 56.77 & 53.33\\
\texttt{B-Arg2} & 63.57 & 61.13 & 62.33\\
\texttt{I-Arg1} & 59.68 & 69.55 & 64.24\\
\texttt{I-Arg2} & 73.08 & 66.74 & 69.77\\
\texttt{O} & 88.86 & 96.51 & 92.53\\
\hline
\end{tabular}
\caption{
Results of the LSTM model on test set using BERT word embedding and parse tree feature. 
}
\label{tab:lstm_bert_tree}
\end{table}

\begin{table}[ht]
\centering
\begin{tabular}{|c|c|c|c|}
\hline
{Label} & {Precision} & {Recall} & {$F_1$}\\
\hline
\texttt{B-Arg1} & 56.33 & 56.86 & 56.59\\
\texttt{B-Arg2} & 65.71 & 64.60 & 65.15\\
\texttt{I-Arg1} & 70.57 & 64.10 & 67.18\\
\texttt{I-Arg2} & 73.90 & 65.96 & 69.70\\
\texttt{O} & 88.72 & 97.68 & 92.98\\
\hline
\end{tabular}
\caption{
Results on test set using fine-tuning BERT model without parse tree features. 
}
\label{tab:bert_finetune}
\end{table}

\begin{table}[ht]
\centering
\begin{tabular}{|c|c|c|c|}
\hline
{Label} & {Precision} & {Recall} & {$F_1$}\\
\hline
\texttt{B-Arg1} & 52.65 & 58.74 & 55.53\\
\texttt{B-Arg2} & 61.27 & 66.10 & 63.60\\
\texttt{I-Arg1} & 63.52 & 70.54 & 66.85\\
\texttt{I-Arg2} & 69.32 & 71.93 & 70.60\\
\texttt{O} & 89.10 & 96.46 & 92.64\\
\hline
\end{tabular}
\caption{
Results on test set using fine-tuning BERT model with parse tree features. 
}
\label{tab:bert_finetune_tree}
\end{table}


\subsection{Additional Results for Sense Classification}
\label{appendix:sense_result}
In this section, we provide the full breakdown of performance of our sense classifier on each sense label. The results are provided in Table~\ref{tab:sense_classification_result_full}. Note that we only included senses existing in the test set. We provide a confusion matrix in Figure~\ref{fig:confusion_matrix}. In addition, we also provide results using 10-fold cross-validation in Table~\ref{tab:sense_classification_result_full_cross_validation}. We also provide the full comparison for sense classification results using predicted arguments versus using gold arguments in Table~\ref{tab:sense_classification_result_pred_vs_gold_full}. We can observe that for most senses, results using predicted arguments have slight inferior performance than those using gold arguments. For cases where results using predicted arguments are better, they are all under-represented senses in the dataset.

\begin{table*}[ht]
\centering
\setlength\tabcolsep{4.5pt}
\begin{tabular}{|l|ccc|}
\hline
Sense & Precision & Recall & $F_{1}$ \\
\hline
Comparison.Concession      & 10.0 & 7.14 & 8.33 \\
Comparison.Contrast      &  40.74 & 84.62 & 55.0 \\
\hline
Contingency.Cause   &   79.43 & 76.45 &  77.91\\
Contingency.Cause+Belief   &  15.0 & 50.0 & 23.08        \\
Contingency.Condition    &    68.52 & 84.09 & 75.51       \\
Contingency.Purpose   &  93.09 & 92.71 & 92.90 \\
\hline
Expansion.Conjunction  &  88.79 & 76.30 & 82.07  \\
Expansion.Disjunction   &  0.00 & 0.00 & 0.00     \\
Expansion.Equivalence     &  0.00 & 0.00 & 0.00   \\
Expansion.Instantiation   &   64.71 & 73.33 & 68.75  \\
Expansion.Level-of-detail  &  60.95 & 53.78 & 57.14    \\
Expansion.Manner         &  48.28 & 68.29 & 56.57     \\
Expansion.Substitution    &   53.33 & 72.73 & 61.54        \\
\hline
Temporal.Asynchronous    &   72.97 & 65.85 & 69.23        \\
Temporal.Synchronous     &   66.67 & 76.92 & 71.43        \\
\hline
Micro Average & 75.19 & 75.19 & 75.19 \\
Weighted Average & 75.98 & 75.19 & 75.19 \\
\hline
\end{tabular}
\caption{Full results on test set for sense classification. We only included senses existing in the test set.}
\label{tab:sense_classification_result_full}
\end{table*}

\begin{table*}[ht]
\centering
\setlength\tabcolsep{4.5pt}
\begin{tabular}{|l|ccc|}
\hline
Sense & Precision & Recall & $F_{1}$ \\
\hline
Comparison.Concession      & 21.75	$\pm$ 21.91 &	17.62 $\pm$	19.07 &	17.15 $\pm$	15.73 \\
Comparison.Contrast      &  45.79 $\pm$	16.14 &	47.23 $\pm$	12.42 &	44.93 $\pm$	12.39 \\
\hline
Contingency.Cause   &   76.50 $\pm$ 4.22 &	74.60 $\pm$ 4.89 &	75.31 $\pm$ 2.14 \\
Contingency.Cause+Belief   &  11.44 $\pm$	8.82 &	21.78 $\pm$	17.88 &	14.62 $\pm$	11.41        \\
Contingency.Condition    &    73.29 $\pm$	9.99 &	77.34 $\pm$	8.69 &	74.28 $\pm$	3.60       \\
Contingency.Purpose   &  86.78 $\pm$ 2.05 &	90.49 $\pm$	2.66 &	88.56 $\pm$	1.57 \\
\hline
Expansion.Conjunction  &  73.99 $\pm$ 8.05 &	70.33 $\pm$	5.55 &	71.82 $\pm$	5.03   \\
Expansion.Disjunction   &  0.00 $\pm$ 0.00 & 0.00 $\pm$ 0.00 & 0.00 $\pm$ 0.00    \\
Expansion.Equivalence     &  24.83 $\pm$	31.10 &	20.67 $\pm$	21.33 &	20.67 $\pm$	21.33   \\
Expansion.Instantiation   &   25.97 $\pm$	14.89 &	39.10 $\pm$	24.73 &	29.11 $\pm$	13.85  \\
Expansion.Level-of-detail  &  57.79	$\pm$ 4.70 &	48.23 $\pm$	3.12 &	52.37 $\pm$	2.06     \\
Expansion.Manner         &  44.10 $\pm$	12.40 &	47.07 $\pm$	9.91 &	44.18 $\pm$	7.96     \\
Expansion.Substitution    &   68.01 $\pm$	22.64 &	66.02 $\pm$	15.04 &	65.23 $\pm$	17.92        \\
\hline
Temporal.Asynchronous    &   57.88 $\pm$	13.06 &	51.39 $\pm$	13.06 &	52.81 $\pm$	9.81        \\
Temporal.Synchronous     &   56.99 $\pm$	13.29 &	53.73 $\pm$	17.83 &	53.84 $\pm$	13.07        \\
\hline
Micro Average & 69.54 $\pm$	1.30 &	69.54 $\pm$	1.30 & 	69.54 $\pm$	1.30 \\
Weighted Average & 70.86 $\pm$	1.45 &	69.54 $\pm$	1.30 &	69.65 $\pm$	1.36 \\
\hline
\end{tabular}
\caption{Full results on cross-validation for sense classification. }
\label{tab:sense_classification_result_full_cross_validation}
\end{table*}

{
\begin{figure*}[t]
\centering
\includegraphics[scale=0.9]{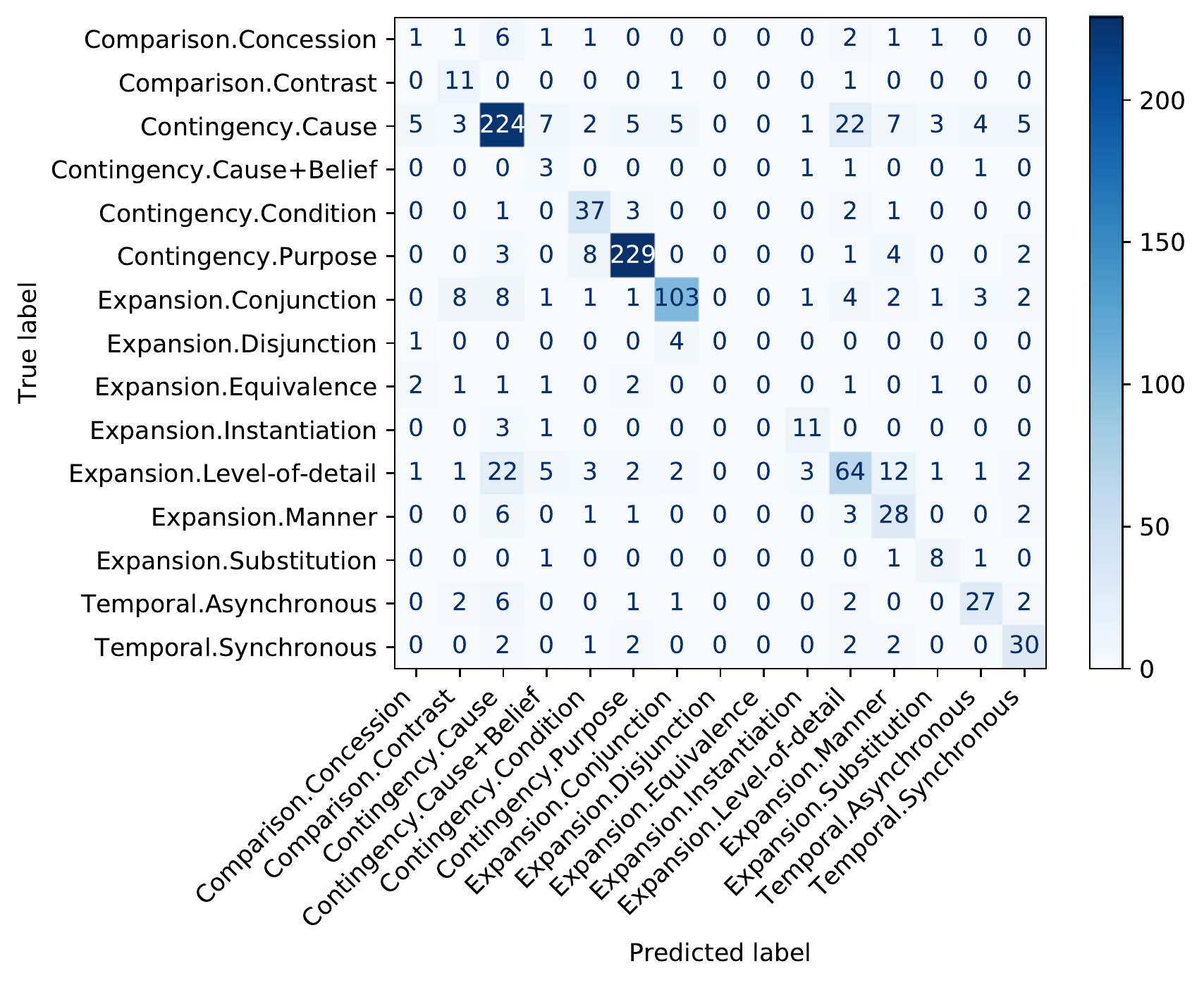}
\caption{\label{fig:confusion_matrix} Confusion matrix of our sense classifier on the test set.}
\end{figure*}
}

\begin{table*}[ht]
\centering
\setlength\tabcolsep{4.5pt}
\begin{tabular}{|l|ccc|ccc|}
\hline
\multicolumn{1}{|c|}{\multirow{2}{*}{Sense}} & \multicolumn{3}{c|}{Predicted arguments} & \multicolumn{3}{c|}{Gold arguments} \\
 & Precision & Recall & $F_{1}$  & Precision & Recall & $F_{1}$ \\
\hline
Comparison.Concession      & 11.11 & 8.33 & 9.52 & 10.0 & 8.33 & 9.09 \\
Comparison.Contrast      &  34.62 & 75.0 & 47.37 & 40.74 & 91.67 & 56.41 \\
\hline
Contingency.Cause   & 75.93 & 73.21 & 74.55 &  79.25 & 75.00 & 77.06\\
Contingency.Cause+Belief   &  17.65 & 60.0 & 27.27 & 11.11 & 40.0 & 17.39       \\
Contingency.Condition    &    66.04 & 85.37 & 74.47 & 62.96 & 82.93 & 71.58       \\
Contingency.Purpose   & 85.90 & 88.64 & 87.25 &  87.01 & 91.36 & 89.14 \\
\hline
Expansion.Conjunction  & 74.55 & 68.33 & 71.30 & 83.00 & 69.17 & 75.45  \\
Expansion.Disjunction   &  0.00 & 0.00 & 0.00 &  0.00 & 0.00 & 0.00    \\
Expansion.Equivalence     &  0.00 & 0.00 & 0.00 &  0.00 & 0.00 & 0.00  \\
Expansion.Instantiation   &  44.44 & 33.33 & 38.10 & 58.33 & 58.33 & 58.33  \\
Expansion.Level-of-detail  & 53.41 & 43.52 & 47.96 & 57.30 & 47.22 & 51.78    \\
Expansion.Manner         &  40.00 & 55.56 & 46.51 & 42.31 & 61.11 & 50.0     \\
Expansion.Substitution    &   57.14 & 72.73 & 64.0 & 50.0 & 72.73 & 59.26        \\
\hline
Temporal.Asynchronous    &   62.86 & 56.41 & 59.46 & 66.67 & 56.41 & 61.11        \\
Temporal.Synchronous     &   65.0  & 68.42 & 66.67 & 63.41 & 68.42 & 65.82        \\
\hline
Micro Average & 69.30 & 69.30 & 69.30 & 71.52 & 71.52 & 71.52 \\
Weighted Average & 69.35 & 69.30 & 68.95 & 72.27 & 71.52 & 71.39 \\
\hline
\end{tabular}
\caption{Comparison of full results on test set for sense classification using predicted arguments versus using gold arguments.}
\label{tab:sense_classification_result_pred_vs_gold_full}
\end{table*}

\end{document}